\documentclass{article}
\usepackage{spconf,amsmath,graphicx}
\usepackage{graphicx}
\usepackage{textcomp}

\usepackage{algorithm}
\usepackage{algpseudocode}
\usepackage{multirow}
\usepackage{xcolor}

\usepackage{pgfplotstable}
\usepackage{colortbl}
\usepackage{xcolor}

\usepackage[inline]{enumitem}

\usepackage[textwidth=15mm]{todonotes}
\usepackage{stfloats}

\algnewcommand\algorithmicforeach{\textbf{for each}}
\algdef{S}[FOR]{ForEach}[1]{\algorithmicforeach\ #1\ \algorithmicdo}
\newcommand{\grayline}{\arrayrulecolor[gray]{0.7}\hline\arrayrulecolor{black}}

\definecolor{myred}{rgb}{0.972, 0.254, 0.266}
\definecolor{myyellow}{rgb}{0.972, 0.777, 0.309}
\definecolor{mygreen}{rgb}{0.262, 0.664, 0.543}
\definecolor{myblue}{rgb}{0.539, 0.535, 0.750}

\title{PLESS: Pseudo-Label  Enhancement with Spreading Scribbles\\for Weakly Supervised Segmentation}
\name{Yeva Gabrielyan$^1$, Varduhi Yeghiazaryan$^1$, and Irina Voiculescu$^2$ 
\thanks{This work was supported by the Afeyan Family Foundation Seed Grants and the JACE Foundation Research Innovation Grant Program at AUA.}}
\address{$^1$Akian College of Science and Engineering, American University of Armenia, Yerevan, Armenia\\
$^2$Department of Computer Science, University of Oxford, Oxford, UK}
\begin{document}
%
\maketitle

\begin{figure*}[!bht]
    \centering
    \includegraphics[width=.9\textwidth]{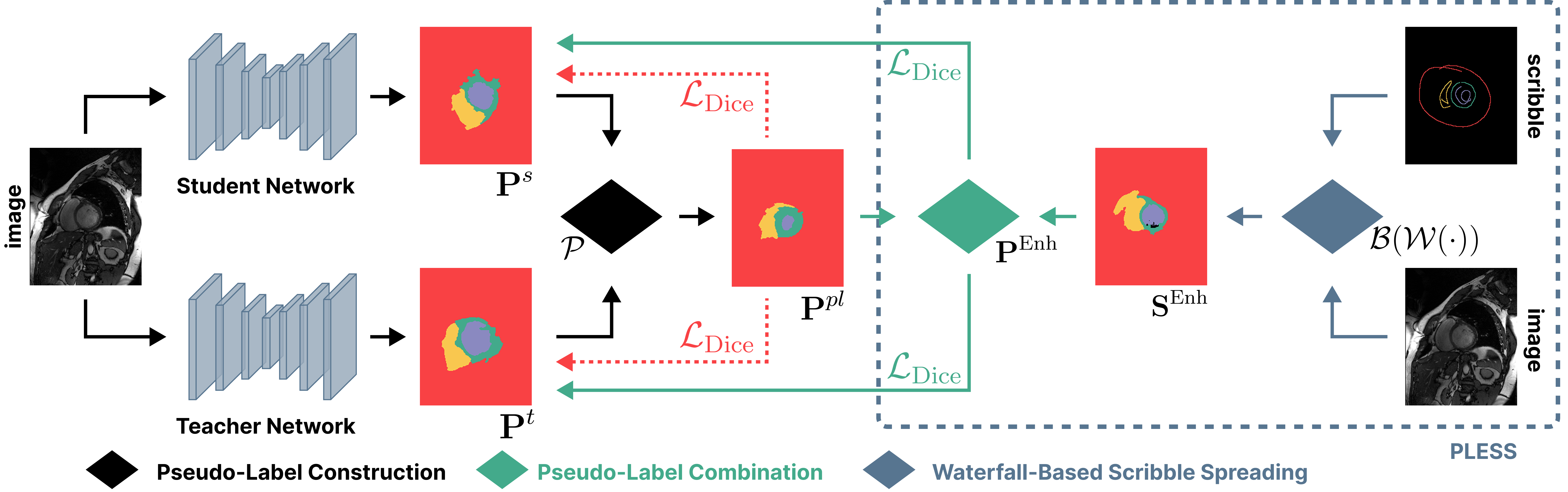}
    \caption{Overview of the proposed PLESS framework, showing pseudo-label construction, waterfall-based enhancement, and integration into pseudo-label loss. The red dotted arrows indicate the original pseudo-label DSC loss formulation used prior to PLESS, which is replaced by the green-arrow loss terms introduced by the proposed enhancement strategy.} 
    \label{fig:PLESS}
\end{figure*}
\begin{abstract}
Weakly supervised learning with scribble annotations uses sparse user-drawn strokes to indicate segmentation labels on a small subset of pixels. This annotation reduces the cost of dense pixel-wise labeling, but  suffers inherently from noisy and incomplete supervision. Recent scribble-based  approaches  in medical image segmentation  address this limitation using pseudo-label-based training; however, the quality of the pseudo-labels remains a key performance limit.
We propose PLESS, a generic pseudo-label  enhancement strategy which improves reliability and spatial consistency. It builds on a hierarchical partitioning of the image into a hierarchy of spatially coherent regions. PLESS propagates scribble information to refine pseudo-labels within semantically coherent regions. 
The framework is model-agnostic and easily integrates into existing pseudo-label methods.
Experiments on two public cardiac MRI datasets (ACDC and MSCMRseg) across four scribble-supervised algorithms show consistent improvements in segmentation accuracy. Code will be made available on GitHub upon acceptance.

\end{abstract}
\section{Introduction}
\label{sec:intro}

To reduce the cost of dense pixel-wise annotation, the segmentation community has increasingly turned to weakly supervised learning methods.
These have been  explored using sparse forms of annotation, including image-level labels, bounding boxes, key points, and scribbles~\cite{pathak2014fully,papandreou2015weakly,bearman2016s,lin2016scribblesup}. Among these, scribbles offer a balance between annotation efficiency and segmentation quality, and have also gained  traction in medical image analysis~\cite{lin2016scribblesup,valvano2021learning,koch2017multi}.

Existing scribble-supervised methods expand sparse annotations using graphical-model-based models~\cite{lin2016scribblesup} or regularization strategies as loss functions~\cite{tang2018regularized}, and more recent approaches leverage adversarial learning or teacher--student frameworks to improve structural consistency~\cite{lin2016scribblesup,valvano2021learning,tang2018regularized,can2018learning,liu2022weakly}. Mixup augmentation methods such as CycleMix~\cite{zhang2022cyclemix} generate blended images, enhancing the training goal with consistency losses to improve segmentation reliability.

\textbf{Pseudo-label-based methods.} Pseudo-labels can be  used to provide supervision for unlabeled images or pixels. During training, pseudo-labels can be generated from either
model predictions or weak annotations, and
treated as supervisory signals. The central idea is to transform imperfect or sparse annotations into higher-quality and more reliable training targets, enabling effective learning from limited supervision~\cite{chen2021semi}. Recent studies have shown that semi-supervised and weakly supervised learning can benefit significantly from high-quality pseudo-labels~\cite{luo2020semi,wu2021semi}.

\textbf{Contributions.} This work proposes a generic pseudo-label enhancement with spreading scribbles (PLESS) to improve the effectiveness of pseudo-label-based training. It  is designed as a model-agnostic framework that can be integrated with existing pseudo-label losses to strengthen supervision quality. The approach is evaluated on four representative scribble-supervised algorithms: DMPLS~\cite{luo2022scribble}, DCDPL~\cite{wang2023weakly}, ScribbleVC~\cite{li2023scribblevc}, and ScribbleVS~\cite{wang2024scribblevs}. 
Our method builds on a hierarchical image partitioning, obtained via watershed and waterfall transforms~\cite{yeghiazaryan2025parallel}, that partitions the image into a hierarchy of regions and enables structured propagation of scribble information. We show how this structure can be further exploited to enhance pseudo-label quality in weakly supervised segmentation.
Comprehensive experiments on two cardiac MRI datasets demonstrate that PLESS consistently improves pseudo-label-based methods, and that ScribbleVS enhanced with PLESS outperforms state-of-the-art (SOTA) approaches. Finally, extensive ablation studies analyze the impact of different components and configurations of PLESS.

\section{Pseudo-label Enhancement}

\begin{table*}[!b]
\caption{Comparison of PLESS against the baseline for the ACDC dataset, using DSC, HD95 (mm) and ASD (mm). Bold indicates the best result for each method, and underlining denotes the overall best.}
\label{tab:acdc_results}
    \centering
    \footnotesize
    \begin{tabular}{l|cccc|c|c||cccc|c|c}
        & \multicolumn{6}{c||}{\textbf{Original}} & \multicolumn{6}{c}{\textbf{PLESS}} \\
        \hline
        & \multicolumn{4}{c|}{\textbf{DSC}} & \textbf{HD95} & \textbf{ASD} & \multicolumn{4}{c|}{\textbf{DSC}} & \textbf{HD95} & \textbf{ASD} \\
        \hline
        & LV & RV & MC & Avg & Avg & Avg & LV & RV & MC & Avg & Avg & Avg \\
        \hline
        \multirow{2}{2.3cm}{DMPLS~\cite{luo2022scribble}} 
        & .881 & .847 & .914 & .881 & 9.37 & 2.53 
        & .886 & .851 & .923 & \textbf{.887} & \textbf{7.10} & \textbf{2.01} \\
        & {\scriptsize {$\pm .002$}} & {\scriptsize {$\pm .005$}} & {\scriptsize {$\pm .006$}} & {\scriptsize {$\pm .003$}} 
        & {\scriptsize {$\pm 2.94$}} & {\scriptsize {$\pm 0.53$}} 
        & {\scriptsize {$\pm .004$}} & {\scriptsize {$\pm .005$}} & {\scriptsize {$\pm .005$}} & {\scriptsize {$\pm .004$}} 
        & {\scriptsize {$\pm 1.63$}} & {\scriptsize {$\pm 0.53$}} \\
        \grayline
        \multirow{2}{2.3cm}{DCDPL~\cite{wang2023weakly}} 
        & .889 & .857 & .916 & \textbf{.887} & 8.36 & 2.45
        & .863 & .823 & .898 & .861 & \textbf{6.26} & \textbf{1.73} \\
        & {\scriptsize {$\pm .003$}} & {\scriptsize {$\pm .007$}} & {\scriptsize {$\pm .004$}} & {\scriptsize {$\pm .004$}} 
        & {\scriptsize {$\pm 1.83$}} & {\scriptsize {$\pm 0.8$}} 
        & {\scriptsize {$\pm .011$}} & {\scriptsize {$\pm .006$}} & {\scriptsize {$\pm .010$}} & {\scriptsize {$\pm .008$}} 
        & {\scriptsize {$\pm 1.22$}} & {\scriptsize {$\pm 0.37$}} \\
        \grayline
        \multirow{2}{2.3cm}{ScribbleVC~\cite{li2023scribblevc}} 
        & .838 & .839 & .904 & .860 & 5.94 & 1.42 
        & .858 & .845 & .907 & \textbf{.870} & \underline{\textbf{4.95}} & \underline{\textbf{1.16}} \\
        & {\scriptsize {$\pm .025$}} & {\scriptsize {$\pm .006$}} & {\scriptsize {$\pm .010$}} & {\scriptsize {$\pm .009$}} 
        & {\scriptsize {$\pm 0.57$}} & {\scriptsize {$\pm 0.21$}} 
        & {\scriptsize {$\pm .012$}} & {\scriptsize {$\pm .007$}} & {\scriptsize {$\pm .012$}} & {\scriptsize {$\pm .007$}} 
        & {\scriptsize {$\pm 1.01$}} & {\scriptsize {$\pm 0.27$}} \\
        \grayline
        \multirow{2}{2.3cm}{ScribbleVS~\cite{wang2024scribblevs}} 
        & .881 & .854 & .917 & .884 & 9.77 & 2.71 
        & .878 & .861 & .925 & \underline{\textbf{.888}} & \textbf{6.39} & \textbf{1.66} \\
        & {\scriptsize {$\pm .015$}} & {\scriptsize {$\pm .016$}} & {\scriptsize {$\pm .014$}} & {\scriptsize {$\pm .015$}} 
        & {\scriptsize {$\pm 6.86$}} & {\scriptsize {$\pm 1.92$}} 
        & {\scriptsize {$\pm .014$}} & {\scriptsize {$\pm .011$}} & {\scriptsize {$\pm .008$}} & {\scriptsize {$\pm .009$}} 
        & {\scriptsize {$\pm 3.13$}} & {\scriptsize {$\pm 0.83$}} \\
    \end{tabular}
\end{table*}

Following~\cite{gabrielyan2025semantic}, a waterfall-based scribble spreading strategy is used to construct spatially enhanced labels. The original approach applies scribble spreading directly to the input scribble annotations to generate training supervision. In contrast, PLESS applies waterfall-based scribble spreading to the constructed pseudo-labels during training.

Each 2D slice is partitioned into a hierarchy of semantically coherent regions. Label classes can denote different anatomical structures of interest, or the background. The labels are then spread inside the coarse layers, then inside the finer layers of the partition hierarchy, as described in~\cite{gabrielyan2025semantic}. In each partitioning layer, regions containing a unique class candidate are assigned that label; regions with conflicting class assignments remain unlabeled. This hierarchical procedure  propagates labels within homogeneous regions while avoiding propagation (`leaks') across ambiguous boundaries.

In order to avoid further label leakage into the background, we introduce an explicit background expansion step to refine the enhanced labels. This step identifies isolated unlabeled regions that are entirely surrounded by background or unlabeled pixels and assigns them to the background label; it also preserves unlabeled regions that touch foreground label boundaries. This operation removes isolated gaps and holes near the image borders and improves the spatial consistency of the enhanced supervision.

The overall PLESS framework is illustrated in Fig.~\ref{fig:PLESS}, where pseudo-labels derived from student and teacher predictions are refined by the PLESS framework via waterfall-based spreading of the scribbles. 

\begin{table*}
\caption{Comparison of PLESS against the baseline for the MSCMRseg dataset, using DSC, HD95 (mm) and ASD (mm). Bold indicates the best result for each method, and underlining denotes the overall best.}
\label{tab:mscmrseg_results}
    \centering
    \footnotesize
    \begin{tabular}{l|cccc|c|c||cccc|c|c}
        & \multicolumn{6}{c||}{\textbf{Original}} & \multicolumn{6}{c}{\textbf{PLESS}} \\
        \hline
        & \multicolumn{4}{c|}{\textbf{DSC}} & \textbf{HD95} & \textbf{ASD} & \multicolumn{4}{c|}{\textbf{DSC}} & \textbf{HD95} & \textbf{ASD} \\
        \hline
        & LV & RV & MC & Avg & Avg & Avg & LV & RV & MC & Avg & Avg & Avg \\
        \hline
        \multirow{2}{2.3cm}{DMPLS~\cite{luo2022scribble}} 
        & .888 & .845 & .924 & \underline{\textbf{.886}} & 8.81 & 2.69 
        & .881 & .841 & .925 & .882 & \textbf{5.40} & \textbf{1.67} \\
        & {\scriptsize {$\pm .003$}} & {\scriptsize {$\pm .003$}} & {\scriptsize {$\pm .003$}} & {\scriptsize {$\pm .002$}} 
        & {\scriptsize {$\pm 0.99$}} & {\scriptsize {$\pm 0.60$}} 
        & {\scriptsize {$\pm .003$}} & {\scriptsize {$\pm .002$}} & {\scriptsize {$\pm .002$}} & {\scriptsize {$\pm .001$}} 
        & {\scriptsize {$\pm 0.69$}} & {\scriptsize {$\pm 0.46$}} \\
        \grayline
        \multirow{2}{2.3cm}{DCDPL~\cite{wang2023weakly}} 
        & .853 & .814 & .921 & \textbf{.862} & 8.21 & 2.78 
        & .818 & .764 & .891 & .825 & \textbf{7.07} & \textbf{2.05} \\
        & {\scriptsize {$\pm .008$}} & {\scriptsize {$\pm .004$}} & {\scriptsize {$\pm .003$}} & {\scriptsize {$\pm .002$}} 
        & {\scriptsize {$\pm 2.52$}} & {\scriptsize {$\pm 0.49$}} 
        & {\scriptsize {$\pm .028$}} & {\scriptsize {$\pm .034$}} & {\scriptsize {$\pm .010$}} & {\scriptsize {$\pm .024$}} 
        & {\scriptsize {$\pm 1.53$}} & {\scriptsize {$\pm 0.47$}} \\
        \grayline
        \multirow{2}{2.3cm}{ScribbleVC~\cite{li2023scribblevc}} 
        & .822 & .812 & .919 & \textbf{.851} & \textbf{6.58} & \textbf{1.69} 
        & .757 & .747 & .912 & .805 & 7.54 & 1.71 \\
        & {\scriptsize {$\pm .051$}} & {\scriptsize {$\pm .017$}} & {\scriptsize {$\pm .003$}} & {\scriptsize {$\pm .022$}} 
        & {\scriptsize {$\pm 0.95$}} & {\scriptsize {$\pm 0.20$}} 
        & {\scriptsize {$\pm .031$}} & {\scriptsize {$\pm .050$}} & {\scriptsize {$\pm .011$}} & {\scriptsize {$\pm .029$}} 
        & {\scriptsize {$\pm 1.62$}} & {\scriptsize {$\pm 0.32$}} \\
        \grayline
        \multirow{2}{2.3cm}{ScribbleVS~\cite{wang2024scribblevs}} 
        & .876 & .844 & .925 & \textbf{.882} & 7.61 & 2.07 
        & .871 & .848 & .928 & \textbf{.882} & \underline{\textbf{5.53}} & \underline{\textbf{1.33}} \\
        & {\scriptsize {$\pm .007$}} & {\scriptsize {$\pm .002$}} & {\scriptsize {$\pm .004$}} & {\scriptsize {$\pm .003$}} 
        & {\scriptsize {$\pm 4.35$}} & {\scriptsize {$\pm 1.16$}} 
        & {\scriptsize {$\pm .007$}} & {\scriptsize {$\pm .005$}} & {\scriptsize {$\pm .003$}} & {\scriptsize {$\pm .004$}} 
        & {\scriptsize {$\pm 1.09$}} & {\scriptsize {$\pm 0.23$}} \\
    \end{tabular}
\end{table*}

Mathematically, let the pseudo-label construction be defined as
$\mathbf{P}^{\mathrm{pl}} = \mathcal{P}(\mathbf{P}^{\mathrm{s}}, \mathbf{P}^{\mathrm{t}}),$
where $\mathbf{P}^{\mathrm{s}}$ and $\mathbf{P}^{\mathrm{t}}$ denote the outputs of the student and teacher networks, respectively, and $\mathcal{P}(\cdot)$ is a network-specific fusion operator that combines the two predictions. The exact form of $\mathcal{P}$ depends on the underlying method and may involve, for example, additive fusion, confidence-based selection, or some other task- or data-specific strategy defined by each algorithm.

The waterfall-based scribble spreading is defined as
$\mathbf{S}^{\mathrm{w}} {=} \mathcal{W}(\mathbf{S}, \mathbf{I}),$
where $\mathbf{S}$ denotes the original scribble annotations and $\mathbf{I}$ is the input image, and $\mathcal{W}(\cdot)$ denotes the hierarchical waterfall-based spreading operator.
To take the enhancement even further, a background expansion operator is applied:
$\mathbf{S}^{\mathrm{Enh}} {=} \mathcal{B}(\mathbf{S}^{\mathrm{w}}),$
where $\mathcal{B}(\cdot)$ fills isolated unlabeled regions with the background label while preserving uncertain areas near foreground boundaries.

Based on the constructed pseudo-labels $\mathbf{P}^{\mathrm{pl}}$ and the enhanced scribbles $\mathbf{S}^{\mathrm{Enh}}$, PLESS defines an enhanced pseudo-label $\mathbf{P}^{\mathrm{Enh}}$ as
\begin{equation}
\mathbf{P}^{\mathrm{Enh}} =
\begin{cases}
\mathbf{P}^{\mathrm{pl}} {\odot} (1 {-} \mathbf{M})
{+}
\mathbf{S}^{\mathrm{Enh}} {\odot} \mathbf{M}
& \text{if } e {\leq} \tau \, E_{\max} \\
\mathbf{P}^{\mathrm{pl}} 
& \text{otherwise}
\end{cases}
\label{eq:enhancement}
\end{equation}
where $\mathbf{M} = \mathbf{I}[\mathbf{S}^{\mathrm{Enh}}]$ is a binary mask indicating pixels with enhanced scribble labels, $\odot$ denotes element-wise multiplication, $e$ is the current training epoch, $E_{\max}$ is the total number of training epochs, and $\tau$ is a tolerance parameter controlling the fraction of training during which pseudo-label scribble spreading is applied.

When using the enhanced pseudo-labels, the Dice-based pseudo-label loss is defined as
\begin{equation}
\mathcal{L}_{\mathrm{pl}} =
\frac{1}{2}
\left(
\mathcal{L}_{\mathrm{DSC}}(\mathbf{P}^{\mathrm{Enh}}, \mathbf{P}^{\mathrm{s}})
+
\mathcal{L}_{\mathrm{DSC}}(\mathbf{P}^{\mathrm{Enh}}, \mathbf{P}^{\mathrm{t}})
\right).
\end{equation}

\section{Experimental Setup}

\textbf{Dataset.} We evaluate our pseudo-label scribble spreading framework on two scribble-annotated cardiac MRI datasets.
\begin{enumerate*}[label=(\alph*)]
\item The ACDC dataset contains cine-MRI scans from 150 patients. Scribble annotations are available for 100 cases, which are used in five-fold cross-validation with 80 scans for training and 20 for validation. The remaining 50 cases without scribbles are used for testing.
\item The MSCMRseg dataset includes late gadolinium enhancement MRI scans from 45 cardiomyopathy patients. The data are split into 25 training, 5 validation, and 15 test scans.
\end{enumerate*}
All results are averaged over five runs.

Both datasets are annotated for three classes: left ventricle (LV), right ventricle (RV), and myocardium (MYO). All results are reported as averages over these classes.

\begin{table*}[!b]
\caption{Comparison with SOTA on the individual features LV, MYO, RV, and their average, evaluated with the Avg DSC.}
\label{tab:SOTA}
\centering
\footnotesize
\begin{tabular}{l|cccc||cccc}

& \multicolumn{4}{c||}{\textbf{ACDC}} 
& \multicolumn{4}{c}{\textbf{MSCMRseg}} \\
\hline
& LV & MYO & RV & Avg & LV & MYO & RV & Avg \\
\hline
\multirow{2}{2.8cm}{CutMix~\cite{yun2019cutmix}} 
& .641 & .734 & .740 & .705 
& .578 & .622 & .761 & .654 \\
& {\scriptsize {$\pm .359$}} & {\scriptsize {$\pm .144$}} & {\scriptsize {$\pm .216$}} &  {} 
& {\scriptsize {$\pm .063$}} & {\scriptsize {$\pm .121$}} & {\scriptsize {$\pm .105$}} &  {} \\
\grayline
\multirow{2}{2.8cm}{Puzzle Mix~\cite{kim2020puzzle}} 
& .663 & .650 & .559 & .624 
& .061 & .634 & .028 & .241 \\
& {\scriptsize {$\pm .333$}} & {\scriptsize {$\pm .231$}} & {\scriptsize {$\pm .343$}} &  {} 
& {\scriptsize {$\pm .021$}} & {\scriptsize {$\pm .084$}} & {\scriptsize {$\pm .012$}} &  {} \\
\grayline
\multirow{2}{2.8cm}{Co-Mixup~\cite{kim2021co}} 
& .622 & .621 & .702 & .648 
& .356 & .343 & .053 & .251 \\
& {\scriptsize {$\pm .304$}} & {\scriptsize {$\pm .214$}} & {\scriptsize {$\pm .211$}} &  {} 
& {\scriptsize {$\pm .075$}} & {\scriptsize {$\pm .067$}} & {\scriptsize {$\pm .022$}} &  {} \\
\grayline
\multirow{2}{2.8cm}{CycleMix~\cite{zhang2022cyclemix}} 
& \textbf{.883} & .798 & .863 & .848 
& .870 & .739 & .791 & .800 \\
& {\scriptsize {$\pm .095$}} & {\scriptsize {$\pm .075$}} & {\scriptsize {$\pm .073$}} &  {} 
& {\scriptsize {$\pm .061$}} & {\scriptsize {$\pm .049$}} & {\scriptsize {$\pm .072$}} &  {} \\
\hline
\multirow{2}{2.8cm}{ScribbleVS PLESS} 
& .878 & \textbf{.861} & \textbf{.925} & \textbf{.888} 
& \textbf{.871} & \textbf{.848} & \textbf{.928} & \textbf{.882} \\
& {\scriptsize {$\pm .014$}} & {\scriptsize {$\pm .011$}} & {\scriptsize {$\pm .008$}} &  {\scriptsize {$\pm .009$}} 
& {\scriptsize {$\pm .007$}} & {\scriptsize {$\pm .005$}} & {\scriptsize {$\pm .003$}} &  {\scriptsize {$\pm .004$}} \\
\end{tabular}
\end{table*}

\textbf{Evaluation Measures.}
During testing, 2D slice predictions are stacked to form 3D volumes. The model with the best validation performance is used for inference. Segmentation quality is evaluated using 3D Dice (DSC), 95th percentile Hausdorff Distance (HD95), and Average Surface Distance (ASD)~\cite{yeghiazaryan2018family}, measured in mm. Higher DSC and lower HD95 and ASD indicate better performance.

\textbf{Implementation Details.}
We follow the protocol of~\cite{luo2022scribble}. Slices are normalized to $[0,1]$, and standard augmentations (rotation, flipping, noise) are applied. Batch sizes are 12 for DCDPL and ScribbleVS, 6 for DMPLS and ScribbleVC. Models are trained for 60K iterations in PyTorch on NVIDIA RTX 4090 GPUs. Aside from the pseudo-label loss, all other losses, e.g., cross-entropy, remain unchanged.

\section{Results}

\begin{figure*}[!t]
    \centering
    \includegraphics[width=.8\textwidth]{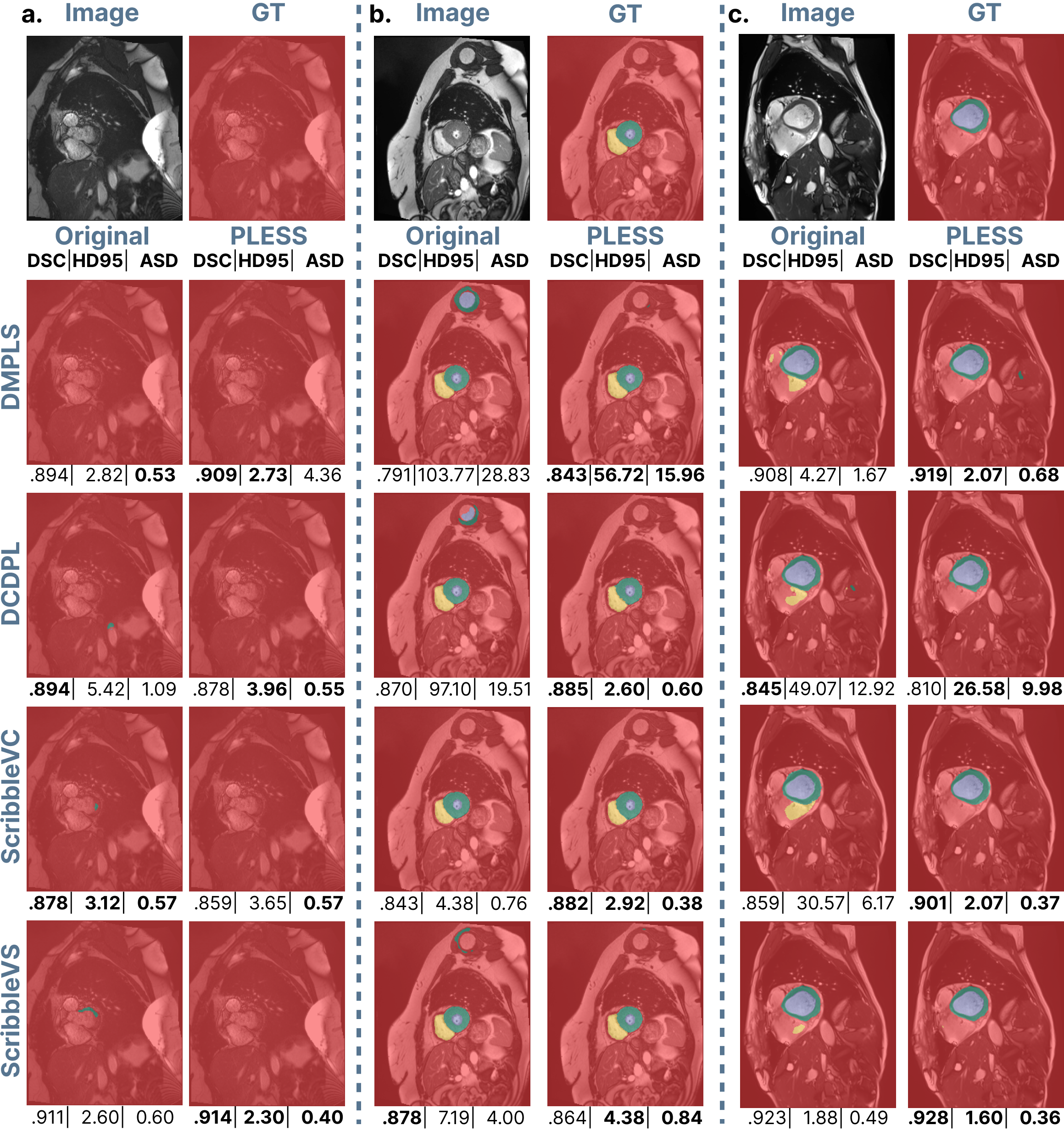}
    \caption{Comparison of segmentation performance using the baseline and PLESS  on three scans (a, b, c) from the ACDC dataset. Row 1 displays the original MRI slice and ground truth (GT), while subsequent rows show segmentation results from different models: DMPLS~\cite{luo2022scribble}, DCDPL~\cite{wang2023weakly}, ScribbleVC~\cite{li2023scribblevc}, and ScribbleVS~\cite{wang2024scribblevs}. Reported DSC, HD95, and ASD scores refer to the whole 3D scan.
    Label colors: background \fcolorbox{black}{myred}{\rule{0pt}{3pt}\rule{3pt}{0pt}}, right ventricle (RV) \fcolorbox{black}{myyellow}{\rule{0pt}{3pt}\rule{3pt}{0pt}}, myocardium (MYO) \fcolorbox{black}{mygreen}{\rule{0pt}{3pt}\rule{3pt}{0pt}}, and left ventricle (LV) \fcolorbox{black}{myblue}{\rule{0pt}{3pt}\rule{3pt}{0pt}}.} 
    
    \label{fig:acdc_qualitative}
\end{figure*}


Performance comparisons between baseline and PLESS are reported in Table~\ref{tab:acdc_results} 
for the ACDC dataset and Table~\ref{tab:mscmrseg_results} for the MSCMRseg dataset. On ACDC, the PLESS setup improves performance across all three measures for all algorithms, except for DCDPL, where improvements are observed for HD95 and ASD only. On MSCMRseg, PLESS leads to consistent improvements in HD95 and ASD for all methods except ScribbleVC. 

For ACDC, the best DSC of .888 is achieved by ScribbleVS with the PLESS setup, while the lowest HD95 of 4.95~mm and ASD of 1.16~mm are obtained by ScribbleVC with PLESS. On MSCMRseg, although the highest DSC of .886 is achieved using the original setup, the best HD95 (5.53~mm) and ASD (1.33~mm) are still obtained with PLESS. 
%
These results demonstrate that PLESS consistently improves boundary accuracy and surface agreement, leading to more reliable segmentations across multiple architectures.

Further comparison with 
SOTA methods that do not employ pseudo-label losses is reported in Table~\ref{tab:SOTA}. On both the ACDC and MSCMRseg datasets, the highest average DSC is achieved by the ScribbleVS algorithm using PLESS. 
It is important to note that the SOTA methods in Table~\ref{tab:SOTA} follow a different experimental protocol, where only the training split of the dataset is used and further divided into separate training, validation, and test sets. In contrast, our experiments follow the standard dataset splits commonly used for fully supervised segmentation, and we use the original training, validation, and test partitions provided by each dataset.

\begin{table*}[!t]
\caption{Ablation study of the different PLESS setups for the ScribbleVS network on the ACDC dataset.}
\label{tab:acdc_ablation}
    \centering
    \footnotesize
    \begin{tabular}{l|ccc|ccc|ccc|ccc}
        & \multicolumn{3}{c|}{\textbf{25\%}} & \multicolumn{3}{c|}{\textbf{50\%}} & \multicolumn{3}{c|}{\textbf{75\%}} & \multicolumn{3}{c}{\textbf{100\%}} \\
        \hline
        & \textbf{DSC} & \textbf{HD95} & \textbf{ASD} & \textbf{DSC} & \textbf{HD95} & \textbf{ASD} & \textbf{DSC} & \textbf{HD95} & \textbf{ASD} & \textbf{DSC} & \textbf{HD95} & \textbf{ASD} \\
        \hline
        \multirow{2}{*}{enh} 
        & \textbf{.888} & 6.68 & 1.73 & .885 & 6.41 & 1.89 & .884 & \textbf{6.33} & 1.88 & .869 & 10.78 & 2.93 \\
        & {\scriptsize {$\pm .012$}} & {\scriptsize {$\pm 3.51$}} & {\scriptsize {$\pm 0.82$}} 
        & {\scriptsize {$\pm .010$}} & {\scriptsize {$\pm 1.02$}} & {\scriptsize {$\pm 0.21$}} 
        & {\scriptsize {$\pm .015$}} & {\scriptsize {$\pm 2.20$}} & {\scriptsize {$\pm 0.62$}} 
        & {\scriptsize {$\pm .007$}} & {\scriptsize {$\pm 3.86$}} & {\scriptsize {$\pm 1.20$}} \\
        \grayline
        \multirow{2}{*}{enh+bg} 
        & \textbf{.888} & 6.39 & \textbf{1.66} & .885 & 8.97 & 2.18 & .886 & 6.42 & 1.71 & .873 & 6.74 & 1.84 \\
        & {\scriptsize {$\pm .009$}} & {\scriptsize {$\pm 3.13$}} & {\scriptsize {$\pm 0.83$}} 
        & {\scriptsize {$\pm .011$}} & {\scriptsize {$\pm 4.64$}} & {\scriptsize {$\pm 1.11$}} 
        & {\scriptsize {$\pm .009$}} & {\scriptsize {$\pm 2.19$}} & {\scriptsize {$\pm 0.63$}} 
        & {\scriptsize {$\pm .011$}} & {\scriptsize {$\pm 2.73$}} & {\scriptsize {$\pm 0.97$}} \\
        \grayline
        \multirow{2}{*}{enh+bg+prop} 
        & \textit{.887} & 6.90 & 1.93 & .886 & 6.96 & 1.93 & .886 & 7.26 & 1.96 & .873 & 8.91 & 2.51 \\
        & {\scriptsize {$\pm .009$}} & {\scriptsize {$\pm 1.69$}} & {\scriptsize {$\pm 0.50$}} 
        & {\scriptsize {$\pm .013$}} & {\scriptsize {$\pm 2.88$}} & {\scriptsize {$\pm 0.98$}} 
        & {\scriptsize {$\pm .013$}} & {\scriptsize {$\pm 2.23$}} & {\scriptsize {$\pm 0.82$}} 
        & {\scriptsize {$\pm .007$}} & {\scriptsize {$\pm 2.22$}} & {\scriptsize {$\pm 0.78$}} \\
    \end{tabular}
\end{table*}

\begin{figure}[!b]
    \centering
    \includegraphics[width=1\linewidth]{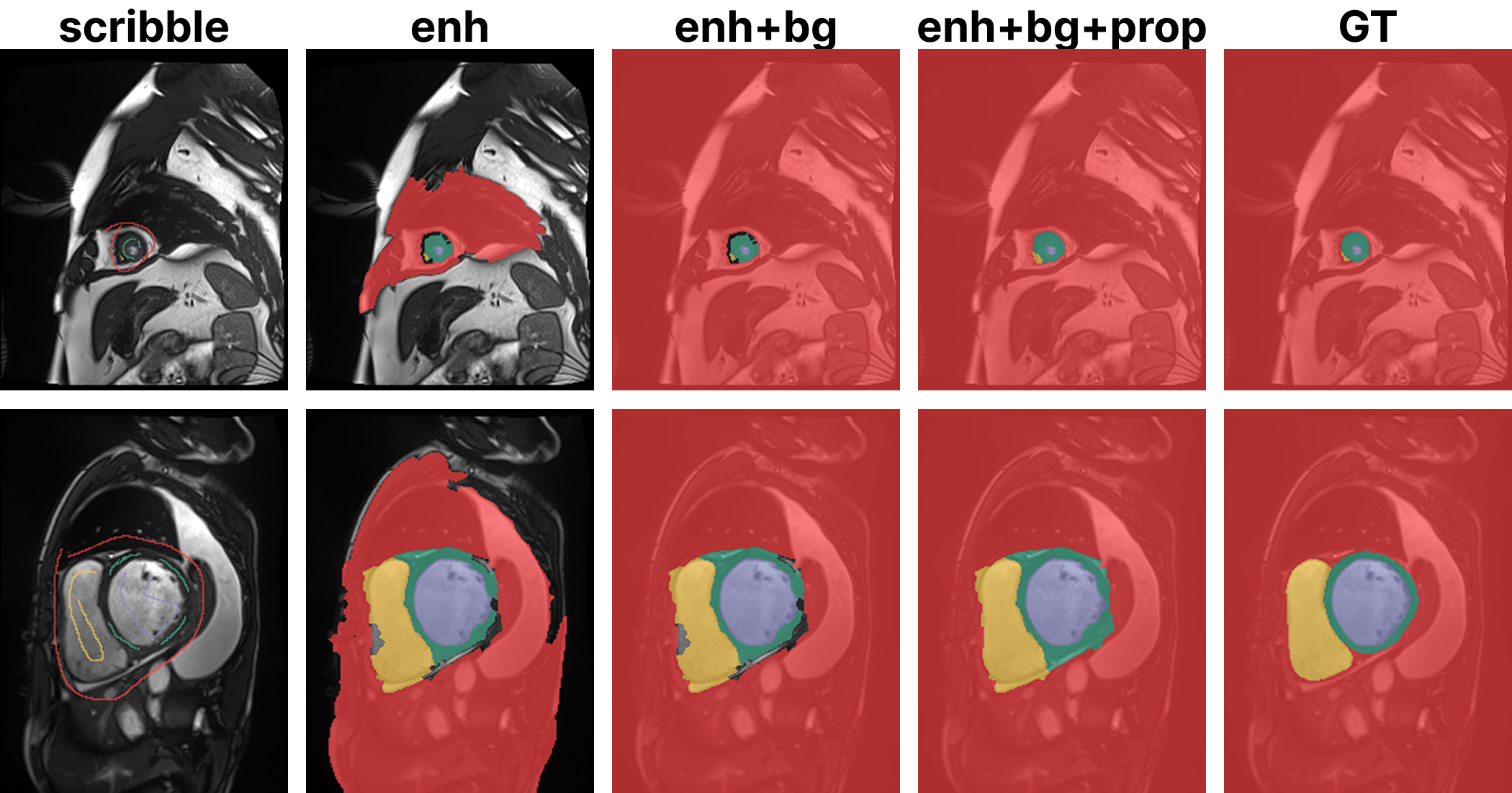}
    \caption{Qualitative comparison among enhancement results with different PLESS setups on a sample from ACDC.} 
    
    \label{fig:enh_viz}
\end{figure}

A qualitative comparison of the performance of all four segmentation methods on the ACDC dataset is presented in Fig.~\ref{fig:acdc_qualitative}. For each image and each method, results for both original and PLESS setups are shown. In most cases, PLESS leads to improvement in boundary delineation, evident by the improved ASD and HD95 scores. 

The three examples in Fig.~\ref{fig:acdc_qualitative} illustrate representative failure cases of the original setup that are mitigated by PLESS. In example (a), for slices where a particular anatomical structure is not present, the original setup for DCDPL, ScribbleVC, and ScribbleVS frequently produces false positive predictions, most notably for the myocardium (MYO). In contrast, with the PLESS setup these errors are fixed.

In example (b), the original setup incorrectly segments regions that visually resemble the target structure, even though they are anatomically unrelated. This is observed for DMPLS, DCDPL, and ScribbleVS, where regions with similar appearance or geometry are mistakenly labeled. The PLESS setup substantially reduces these errors, suggesting that the propagated labels provide stronger spatial and contextual information, which helps the algorithms differentiate between similar-looking structures.

In example (c), when only two of the three cardiac structures are present in a slice, the original setup often hallucinates the missing third class. This behavior is observed across all four methods and reflects a common limitation of weak supervision, where models tend to over-predict plausible anatomical classes. With PLESS, these false positive segmentations are reduced.
These qualitative results show that PLESS improves not only segmentation accuracy, but also the robustness of predictions in challenging and ambiguous cases.

\subsection{Ablation Studies}

To identify the most effective PLESS configuration, an ablation study is conducted on two key components. Different strategies for scribble spreading are first analyzed, considering three progressive variants. The baseline scribble spreading (\textit{enh}) corresponds to the original waterfall-based region expansion proposed in~\cite{gabrielyan2025semantic} and \textit{enh+bg} is the introduced background expansion variant. Finally, full region propagation (\textit{enh+bg+prop}) is applied, where the remaining unlabeled pixels are iteratively assigned labels from neighboring regions until full image coverage is achieved. The three different stages of PLESS are visualized in Fig.~\ref{fig:enh_viz}.

The second key component is the tolerance level. Table~\ref{tab:acdc_ablation} reports the performance of the ScribbleVS network under different PLESS configurations and tolerance levels, ranging from 25\% to 100\%. The \textit{enh+bg} variant achieves the best performance across two of the measures (DSC and ASD). Performance is strongest at 25\% and 75\%, while a degradation is observed at 100\%. This suggests that aggressive scribble spreading at full tolerance can introduce noise, as the enhanced annotations may contain inaccuracies and ambiguities. At later training stages, pseudo-labels generated by the teacher and student networks can be more reliable than the propagated enhancements themselves. This leads to better performance when scribble spreading is not applied in the later episodes. In most cases, the variant without full propagation (\textit{enh+bg}) outperforms \textit{enh+bg+prop}, indicating that complete label propagation introduces incorrect labels in uncertain regions, which degrades performance.

\section{Conclusion}

The PLESS framework exploits spatial structure to improve the reliability of pseudo-label supervision in weakly supervised segmentation. Rather than introducing additional model complexity, PLESS operates at the level of image structure. Since its supervision is fairly generic, it can be integrated into existing training pipelines with minimal overhead. The observed performance gains across multiple scribble-supervised methods suggest that explicitly encoding spatial coherence during pseudo-label refinement is a powerful direction related to traditional edge detection and image segmentation. Beyond scribble-based supervision, the proposed strategy is applicable to other forms of sparse or noisy annotation, making it a flexible tool for a broad range of weakly supervised segmentation settings, including those outside the medical domain.

\bibliographystyle{IEEEbib}
\bibliography{strings,refs}

\end{document}